\documentclass[10pt,twocolumn,letterpaper]{article}

\usepackage{cvpr}
\usepackage{times}
\usepackage{epsfig}
\usepackage{graphicx}
\usepackage{amsmath}
\usepackage{amssymb}

\usepackage[pagebackref=true,breaklinks=true,letterpaper=true,colorlinks,bookmarks=false]{hyperref}

\cvprfinalcopy 

\def\cvprPaperID{1017} 

\ifcvprfinal\fi
\begin{document}

\title{Web-Scale Training for Face Identification}

\author{
Yaniv Taigman, Ming Yang, Marc'Aurelio Ranzato\\
Facebook AI Research\\
Menlo Park, CA 94025, USA \\
\tt{\small \{yaniv, mingyang, ranzato\}@fb.com} \\
\and
Lior Wolf \\
Tel Aviv University \\
Tel Aviv, Israel \\
\tt{\small wolf@cs.tau.ac.il} \\
}

\maketitle

\begin{abstract}
Scaling machine learning methods to very large datasets has attracted considerable attention in recent years, thanks to
easy access to ubiquitous sensing and data from the web.
We study face recognition and show that three distinct properties have surprising effects on the transferability of deep convolutional networks (CNN): (1) The bottleneck of the network serves as an important transfer learning regularizer, and (2) in contrast to the common wisdom, performance saturation may exist in CNN's (as the number of training samples grows); we propose a solution for alleviating this by replacing the naive random subsampling of the training set with a bootstrapping process. Moreover, (3) we find a link between the representation norm and the ability to discriminate in a target domain, which sheds lights on how such networks represent faces.
Based on these discoveries, we are able to improve face recognition accuracy on the widely used LFW benchmark, both in the verification (1:1) and identification (1:N) protocols, and directly compare, for the first time, with the state of the art Commercially-Off-The-Shelf system and show a sizable leap in performance.
\end{abstract}

\section{Introduction}
\label{intro}
Face identification is a recognition task of great practical interest for which (i) much larger labeled datasets exist, containing billions of images; (ii) the number of classes can reach tens of millions or more; and (iii) complex features are necessary in order to encode subtle differences between subjects, while maintaining invariance to factors such as pose, illumination, and aging. 
Performance improvements in recent years have been staggering, and automatic recognition systems cope much better today with the challenges of the field than they did just a few years ago. These challenges are well documented and include changing  illumination, pose and facial expression of the subject, occlusion, variability of 
facial features due to aging, and more. In the past, impressive performance was only demonstrated for
carefully registered face images that did not exhibit much variability, a set up commonly referred to as {\em 
constrained conditions}. 
Current state-of-the-art methods for \emph{unconstrained} face recognition and verification (the task of predicting
whether two face images belong to the same person or not)~\cite{Taigman:CVPR14, Cao:ICCV13, Sun:ICCV13} employ a similar
protocol: they use a fairly large collection of images to learn a robust representation or metric, and then
they perform transfer learning to predict the identity or the similarity between two face images.
These methods are trained on hundreds of thousands or a few million images and recognize up to a few thousand different subjects. This is orders of magnitude larger than what was ever attempted in the past, yet two or three orders of magnitude smaller than the actual datasets available today. 
Our starting point is the DeepFace architecture~\cite{Taigman:CVPR14}, which is based on recent advances in deep learning. 
Our contribution is not only to redefine the state of the art on a public benchmark using an improved system, but also:
(i) we study the role of the bottleneck as a transfer learning regularizer and (ii) we propose a new way to utilize large datasets by replacing the standard random sub sampling procedure with a bootstrapping procedure; 
(iii) we discover a three-way link between the representation norm, the image quality, and the classification confidence. 

These discoveries lead to an improvement in verification performance on a widely used benchmark dataset, the LFW dataset~\cite{Huang:ECCVW08}. We then turn our attention to the 1:N identification problem, in which the image is identified out of a gallery of N persons. While the usage of verification datasets has advanced the field of computer vision greatly, the 1:N scenario is much more directly related to face identification applications. For instance, vendor tests, e.g.,~\cite{Grother:MBE10}, focus on 1:N protocols. As a result, the research community was unable to directly compare academic contributions to the most prominent commercial systems. By making use of a recently proposed 1:N benchmark built on top of the LFW images~\cite{Lacey:MSUTR}, we are able to perform this comparison for the first time. 




\section{Previous work}
\label{previous}

The U.S. National Institute of Standards and Technology (NIST) publicly reports every several years its internal benchmark on face recognition; systems taking part in the competition are developed by leading commercial vendors as well as a few research labs. For instance, in the MBE 2010 report~\cite{Grother:MBE10}, the three top-ranked Commercial Off The Shelf  (COTS) correctly matched probed faces against a large collection (``gallery'' in the commonly used terminology) of 1.6 million identities with an 82\%-92\% accuracy rate.
The datasets used by~\cite{Grother:MBE10,Klontz:Boston13} are not publicly available, and therefore it is hard to compare the performance of academic systems on the same benchmarks. Fortunately, the same COTS system was recently tested~\cite{Lacey:MSUTR} on a 1:N identification benchmark constructed using the images of the public Labeled Faces in the Wild (LFW)~\cite{Huang:ECCVW08} dataset.
On this benchmark, the rank-1 accuracy of the COTS system dropped to about 56\%~\cite{Lacey:MSUTR}, even though the gallery has only a couple of thousands identities. This finding demonstrates that although constrained face recognition has reached an impressive accuracy, the unconstrained one is still far from being solved.

Face {\em verification}, which is the task of determining whether two face images belong to the same subject, has greatly advanced in recent years, especially in the unconstrained setting. In fact, recent contributions~\cite{Sun:CVPR14,Taigman:CVPR14, DBLP:journals/corr/SunWT14} reported nearly human level performance on the LFW verification task using deep neural networks, but no test is reported on the probe-gallery identification task.

Scaling up face recognition is a non-trivial challenge. The baseline DeepFace system~\cite{Taigman:CVPR14} has about 100 million parameters to start with. It is very hard to distribute efficiently~\cite{Hinton:NC06,Krizhevsky:arXiv14}. It produced features that are lower dimensional than engineered features~\cite{Cao:ICCV13}  but still contain several thousand dimensions; and it needs massive amounts of data to generalize well~\cite{Krizhevsky:NIPS12,Szegedy:arXiv13}. There is no known method to effectively train such a large system on billions of images with millions of labels, using thousands of features.
In the machine learning literature, several methods have been proposed to deal with very large datasets. The simplest method is to randomly down-sample the dataset, which is a clearly sub-optimal approach. A more suitable alternative is to employ bootstrapping procedures that aim at focusing on the hardest cases ignoring or down-weighing the easy ones, like in boosting~\cite{JMLR:v15:dubout14a}. The approach we advocate for is, in essence, a bootstrapping one since  we focus the training effort on a cleverly selected subset of the samples that are hard to classify. However, the selection process is made much more efficient than in standard bootstrapping because we do not need to first evaluate each training sample in order to perform our selection. \\
As part of our sample selection process, we utilize the similarity between classes based on the parameters of their classifiers. Similarity of classifiers has been used in the literature in other contexts. In~\cite{malisiewicz-iccv11} multiple SVM classifiers, each based on a single positive sample, are used to construct a powerful descriptor of the learned class. Note that in our case, the final descriptor is trained in one multi-class classification network, whereas SVMs are only used to select labels for training this network.

Our contribution continues a line of work that has gained considerable recent attention -- understanding the underlying mechanisms behind the ``unreasonable'' success of deep networks. In~\cite{Zeiler:ECCV14} the deep activations are propagated back to the image in order to give insights into the role of the intermediate layers and the operation of the classification layer. In~\cite{Szegedy:arXiv13}, optimization is used to trick the CNN to misclassify clear input images and to compute the stability of each layer. Recently,~\cite{bengionips} studied empirically the tradeoff between the generality and specificity of each layer.

These contributions mostly focus on image classification trained on ImageNet, a highly varied dataset with a few million images and 1,000 classes. Faces, since they have a clear structure, training data in abundance, and well understood challenges, provide a unique opportunity for understanding the basic properties of CNN-based transfer learning.

\begin{figure}
\includegraphics[width=0.99\linewidth]{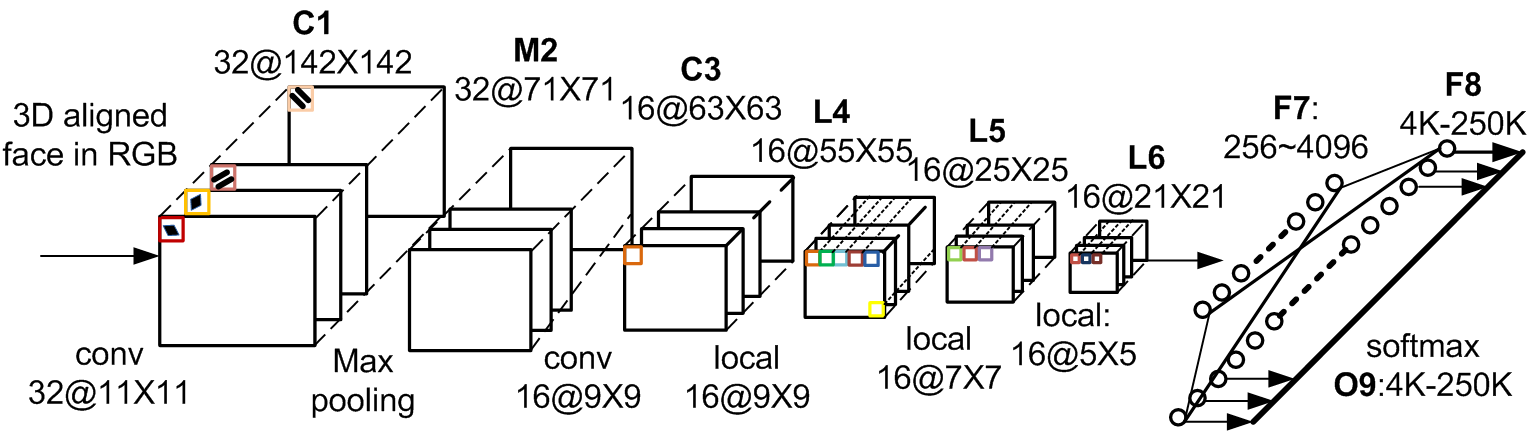}
\caption{\footnotesize \textbf{The initial baseline network architecture.} A front end of convolutional, pooling, convolutional layers is followed by three locally connected layers and two fully connected layers.}
\label{fig:oldarch}
\end{figure}

\section{Transfer Learning in Faces}
\label{proposed}

In this section, we describe our framework, starting with an initial face representation, which was trained similarly to DeepFace~\cite{Taigman:CVPR14}, and exploit discoveries in this network to scale the training up to, the previously unexplored range of, hundreds of millions of training images.



DeepFace was shown to achieve good generalization. However, the association between the transferability of the network to its design was left largely unexplored. We report three properties that strongly affect the quality of the transfer. First, we show that the dimensionality of the representation layer (F7), which we will refer to as \emph{bottleneck}, dramatically affects transferability. Second, we find that selecting random samples from a very large pool of samples leads to performance saturation, which can be alleviated by replacing the naive random subsampling practice with a semantic bootstrapping method. Third, we link measurable properties of the transferred representations to the expected performance at the target domain.


\subsection{Baseline DeepFace Representation}
\label{sec:baseline}
A total of four million face images belonging to 4,030 anonymized identities (classes), were aligned with a 3D model and used for learning an \emph{initial} face representation, based on a deep convolutional neural network. As shown in Fig.~\ref{fig:oldarch}, the network consists of a front-end two convolutional layers with a single max-pooling layer in between (C1-M2-C3), followed by three locally-connected layers L4-L5-L6, without weight sharing, and two fully-connected layers F7-F8. The output of F8 is fed to a 4030-way softmax which produces a distribution over the class labels.  Denote by $o_i(x)$ the $i$-th output of the network on a given input $x$,  the probability assigned to the $i$-th class is the output of the softmax function: $p_i(x) = \exp(o_i(x)) / \sum_j \exp(o_j(x))$. The $ReLU(a) = max(0, a)$ nonlinearity ~\cite{relu} is applied after every layer (except for F8) and optimization is done through stochastic gradient descent and standard back-propagation~\cite{backprop,LeCun:PROC98}, minimizing the cross-entropy loss. If $k$ is the index of the true label for a given input $x$, the loss associated with this sample is: $L(x) = - \log p_k(x)$. Once trained, the representation used is the normalized feature vector of layer F7~\cite{Taigman:CVPR14}.


\subsection{Bottleneck and Transferability}
\label{sec:compress}

\begin{figure}
\centering
\includegraphics[width=0.8\linewidth]{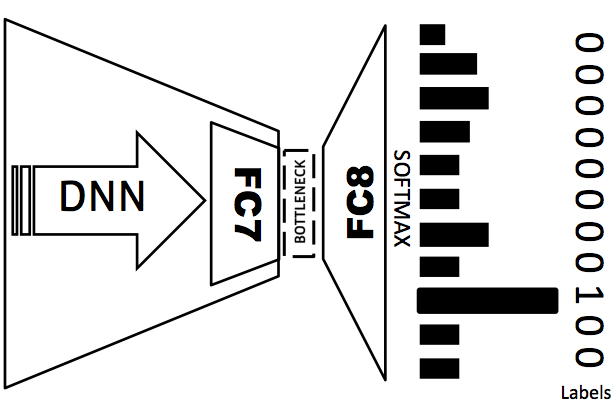}
\caption{\footnotesize \textbf{The bottleneck. } The representation layer splits the network between the part that converts the input into a {\em generic} face descriptor and the part that performs linear classification to {\em specific} K classes. FC7 and FC8 are the low-rank matrices that project to- and from the bottleneck.}
\label{fig:bottle}
\end{figure}
We show that the dimensionality of the last fully-connected layers (F7 \& F8) critically affects the balance between generality and specificity. As illustrated in Fig~\ref{fig:bottle}, in a K-way multiclass network with binary targets, the \emph{classification} layer (F8) is a collection of K linear (dependent) classifiers.

By compressing the preceding \emph{representation} layer (F7) through a lower rank weight matrix, we reduce the ability of the network to encode training-set specific information in this layer, thereby shifting much of the specificity to the subsequent classification layer (F8).
For the purpose of transfer learning, the classification layer is ignored once training finishes, and the network is regarded as a feature extractor. A compact bottleneck, therefore, decreases the network specialization and increases the representation generality.
However, a narrower bottleneck increases the difficulty of optimizing the network when training from scratch. For the architecture described in Sec.~\ref{sec:baseline}, we are able to effectively train with bottlenecks of dimensionality as low as 1024, but not lower.
For smaller dimensions the training error stopped decreasing early on. However, we note a useful property: by pre-loading the weights of the initial network, except for the last layer, we were able to learn much smaller embeddings effectively, as furthered detailed in Sec.~\ref{sec:results}. Remarkably, a bottleneck of only 256 dimensions yielded convincingly better accuracies on the target domain in all configurations.

This might be counter intuitive, since the common wisdom~\cite{Chen:CVPR13} is that the representation should be as large as the number of samples can support. However, as we show, decreasing the representation size is beneficial even if the number of training samples is virtually unlimited. The reason is that we learn our representation in one domain and test it on another, which does not share the same underlying distribution. Therefore, regularization is warranted.
In addition to generalization, such compact representations enable us, efficiency wise, to scale up the experiments by an order of magnitude, exploring much larger configurations as we describe next.

\subsection{Semantic Bootstrapping}

A conjecture is made in~\cite{Krizhevsky:NIPS12} that ``results can be improved simply by waiting for faster GPUs and bigger datasets to become available''. Our findings reveal that this holds only to a certain degree. Unlike the popular ImageNet challenge, where a \emph{closed} collection of 1.6 million images is split into train and test, we can access a much larger dataset in order to test, for the first time, both the transferability and scalability properties of deep convolutional nets at scales three orders of magnitude larger. We find that using the standard Stochastic Gradient Descent (SGD) and back-propagation leads to performance saturation in the target domain when the training set size in the source domain grows beyond a certain point, as shown in Sec.~\ref{sec:results}. This holds even when we change the architecture of network, by either adding more layers and/or increasing their capacity. By judiciously selecting samples as opposed to picking them at random, we were able to improve performance further.

We leverage the compressed representation presented above in order to train efficiently. The dataset at our disposal contains 10 million anonymized subjects with 50 images each in average. This is to be compared to around 4000 identities and 4 million images ($DB_1$) in~\cite{Taigman:CVPR14}. This new dataset was randomly sampled from a social network. By running string matching, we verified that the identities do not intersect those of LFW. 

Having this larger dataset at our disposal, we search for impostor samples to be used in a second round of training, as normally done in bootstrapping. The most basic method would be to sample a large pool of face representations, each represented by the compact feature, and select the nearest neighbors from other identities (``impostors''). However, for reasons specified below, a significant improvement in both scalability and reliability is obtained by working in the space of linear models, trained discriminatively on top of the compressed representations, as opposed to directly working with the compressed features.

As a first step, we represent each class by a single classifier, i.e., for each identity, we learn a hyperplane trained in a binary classification setting of one-vs-all, where the positive instances (representations) are of the same identity and the negatives are a random subset of other identities. Each identity is associated with an average of 50 face images. Therefore, working with linear models, instead of the underlying instances, enables us to scale the exploration of impostor identities by another order of magnitude. In terms of efficiency, training such linear models can be easily parallelized, and it is very efficient especially when using compact features.

In addition to being more scalable, this semantic distance performs better than instance-based bootstrapping, probably due to its added robustness to human labeling errors of the ground-truth.
When sampling pairs of same/not same samples based on instance similarity, we notice that many of the pairs sampled, in particular those that are labeled as the 'same' identity, are of different individuals due to labeling errors. Performing, for example, metric learning on such data leads to reduced performance compared to the baseline representation. Specifically, a Siamese network trained on 4M nearest neighbor pair instances, belonging either to the same class or not (=impostors), obtained substantially worse results than the baseline system. This observation is also consistent with the fact that the initial representation is already on-par with humans w.r.t. pair-wise verification performance, thereby flushing out the ground-truth errors.

Bootstrapping is performed as follows: we randomly select 100 identities, as seeds, among the 10 million models. For each seed, we search for the 1000 nearest models, where the similarity between any two models $h_1, h_2$  is defined as the cosine of the angle between the associated hyperplanes: $S(h_1, h_2) = <h_1, h_2> / (\left \| h_1 \right \| \left \| h_2 \right \|)$. The union of all images of all retrieved identities constitutes the new bootstrapped dataset $DB_2$, containing 55,000 identities overall. Note that in contrast to negatives sampling~\cite{Weston:2011}, as normally done in bootstrapping, $DB_2$ consists of both easy \& hard samples - separating between seeds is as easy as before, but much harder inside the neighborhood of each seed, by construction.

In terms of efficiency, the training of $10^7$ hyperplanes was completed in 2 days utilizing several 256G RAM commodity servers. Evaluating the distance between each seed and a gallery pool of these $10^7$ hyperplanes reduces to a matrix-multiplication $W s_i$ where $W$ is a matrix of $10^{7} \times 256$ and seed $s_i \in R^{256}$ is a single seed. The run time of this step on a single server is about 1 second per seed query. Overall, the entire subset selection process takes an hour on a single high memory server.

\subsection{Final Network Architecture}
\label{sec:32filters}
$DB_2$ is a challenging dataset, and our objective is to train feature representation that can discriminate between the new selected identities.
Increasing the dimensionality of the representation and consequently training a bigger network is, therefore, essential in order to model the subtle differences between lookalike subjects found more abundantly in the bootstrapped training set (see Sec~\ref{sec:results}).

\begin{figure}

\includegraphics[width=0.99\linewidth]{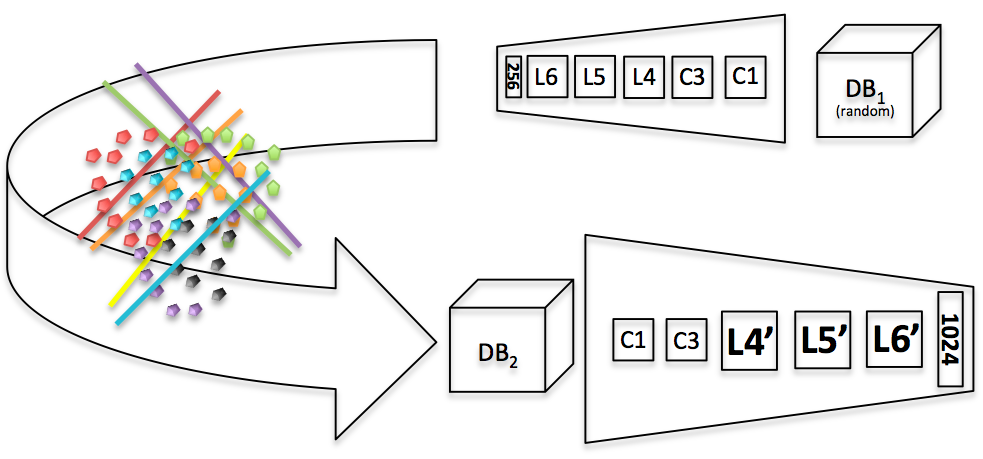}

\caption{\footnotesize \textbf{The bootstrapping method.} An initial 256D-compressed representation trained on $DB_1$ is used to find the semantically-nearest identities of randomly picked 100 seeds, in a large pool of pre-trained hyperplanes. The union of all 100 groups of selected identities define the bootstrapped dataset $DB_2$. A larger capacity network with enlarged locally-connected layers and a 1024D representation is then trained.}
\label{fig:newarch}
\end{figure}

Specifically, we pre-load C1 and C2 layers from the initial network, and double the number of filters of each locally-connected layer from 16 to 32. In addition, we enlarge the representation layer F7 from 256 to 1024. All new layers, except for C1 and C2, are randomly initialized and trained on $DB2$, with the same algorithm as before. The two first convolutional layers C1 and C2 are merely feature extractors, which include less than 1\% of the overall weights. Empirically, we found that fixing these layers did not affect performance while speeding up training considerably.
Unlike boosting and cascaded architectures~\cite{Viola:CVPR01}, we end up
with a single classifier and are interested in learning representations that are useful for transfer learning. Figure ~\ref{fig:newarch} visualizes the process.

\section{The Representation Norm}
\label{sec:norm}

One of the biggest benefits of DNNs is their ability to learn representations that can generalize across datasets and tasks~\cite{bengionips}.
Faces, since they have a clear structure, training data in abundance, and well understood challenges, provide a unique opportunity to discover basic properties of this increasingly popular form of transfer learning.

\begin{figure}[t]
\centering
\includegraphics[width=0.90\linewidth]{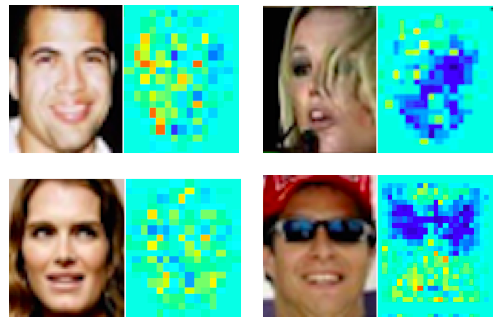}
\caption{Examples of L6 activations for various faces. In each pair the original image is compared to the sum of the channels of its L6 activations: (left) pairs depicting good quality images. (right) examples of poor quality images (occluded and/or misaligned). Bluer is lower, red is higher. Best viewed in color. }
\label{fig:local_relu.png}
\end{figure}

The input faces are \emph{aligned}, and since the spatial structure of our deep architecture is preserved until the representation layer, the feature maps remain a well localized description of the underlying face. Inspecting the topmost local layer (L6) provides easy to decipher information regarding the underlying image. Consider Fig.~\ref{fig:local_relu.png} that shows the original image and its L6 (summing across all feature maps) side by side. Occlusions and other types of local distortions in the input face image lead to weaker activations in the corresponding areas of L6.

A pixel of the input image contributes to a value in a deep layer only if there is a path, from this location in the input layer to the deep layer, for which all activations pass the ReLU thresholds. Paths originating in disrupted regions tend to get blocked, leading to darkened matching regions in the top local layers.

In fact, the information of which unit is activated in the representation holds most of the discriminative information. When applying a simple threshold at zero to the image representation (F7), the resulting binary vector remains highly discriminative. On the LFW benchmark, the performance drop that follows the binarization of the representation is typically only 1\% or less.

\begin{figure*}[t]
\centering
\begin{tabular}{cccc}
\includegraphics[width=0.23\linewidth]{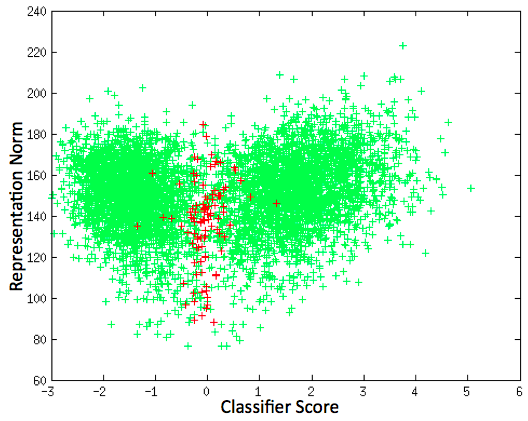} &
 \includegraphics[width=0.23\linewidth]{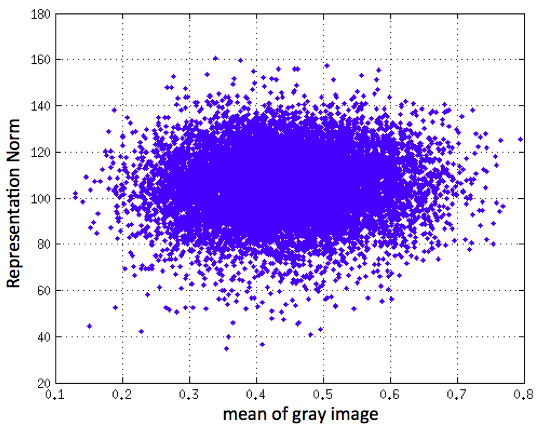} &
 \includegraphics[width=0.23\linewidth]{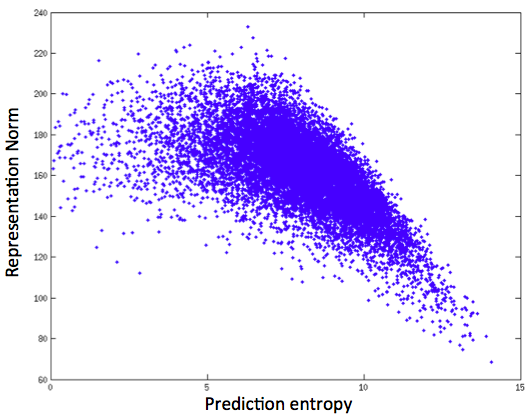} & \includegraphics[width=0.23\linewidth]{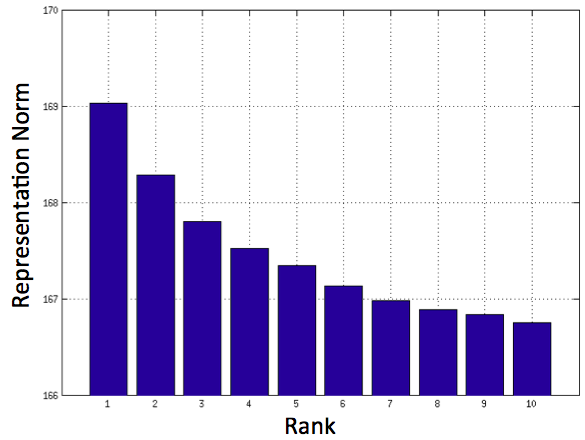}\\
(a) & (b)& (c)& (d)\\
\end{tabular}
\caption{\small Connecting the representation norm to image distinguishability. (a) The classifier score (signed distance from the separating hyperplane) for same not-same prediction on the LFW benchmark vs. $\min(||r_l||,||r_r||)$, where $r_l$ and $r_r$ are the representations of the image pair. Red denotes misclassification. As can be seen, lower confidence predictions and mistakes tend to have lower representation norms. (b) The mean intensity value of the face region vs. the representation norm. High intensity images are not typically linked with higher representation norms. (c) Representation norm vs. prediction entropy. The higher the representation norm is, the lower the uncertainty is expected to be. (d) Retrieval rank vs. mean representation norm on our internal validation set (Sec.~\ref{sec:modelsec}). Misclassified probes (rank$>$1) tend to have a lower norm than correctly matched probes (rank=1).}
\label{fig:norms}
\end{figure*}

Since image disruptions lead to localized L6 inactivity, and since F7 is a linear projection of L6 followed by a threshold, these disruptions lead to a reduced norm of the representation vector F7. This can be seen in Fig.~\ref{fig:norms}(a), in which the link between the norm of a pair of face representations is compared to the certainty of the classifier mapping pairs of images to the same/not-same identity. Curiously enough, while the representation in a ReLU-network is directly linked to the norm of the representation, the link between the image mean intensity value and its representation norm is much weaker, as can be seen, for all LFW images in Fig.~\ref{fig:norms}(b).

A more direct way to explore the link between the representation norm and the image discriminativity is to consider the entropy of the classification layer, i.e., the outputs of the softmax layer. This is shown in Fig.~\ref{fig:norms}(c), where for all LFW images the representation norm is plotted against the entropy of the probabilities obtained by the original deepface network, trained on the 50,000 identities of $DB_2$. This link is very strong and a correlation below -0.65 is obtained between the norm and the entropy.

In order to explain the link between the representation norm and the entropy, we use the first order Taylor expansion of the classifier output.
Let $r$ be the vector of activations of F7.
Let $z$ be the vector of F8 activations obtained for $r$, then, $z=Wr$, where $W$ contains the weights mapping F7 to F8. The softmax function maps $z$ to probabilities: $p_i = \frac{e^{z_i}}{\sum_j e^{z_j}} \approx \frac{1+z_i}{N+\sum_j z_j}$, where the last approximation holds for small $z_i$, since $e^x\approx 1+x$ for small $x$.

The approximation of the entropy is therefore $H(p) \approx -\sum_i \frac{1+z_i}{N+\sum_j z_j} \log \frac{1+z_i}{N+\sum_j z_j}$. Since the softmax vector of $z$ is the same as the softmax vector of $z+b$ for all b, we assume, w.l.o.g, that the mean of $z$ is zero and obtain that $H(p) \approx -\sum_i \frac{1+z_i}{N} \log \frac{1+z_i}{N}$. By the approximation above, $log(1+z_i) \approx z_i$ and we obtain:
$$H(p) \approx -\sum_i \frac{1+z_i}{N} (z_i-\log(N))~~.$$

We now consider a family of scaled version of $r$: $sr$, where $s$ is a scale factor. Since $z=Wr$, the activations of F8 also scale with $s$, and the entropy approximation of the scaled activations become $-\sum_i \frac{1+sz_i}{N} (sz_i-\log(N)) $. This expression is dominated, for small values of $s$, by a linear function of $s$, which explains the behavior seen on the left side of Fig.~\ref{fig:norms}(c).

To conclude, lower representation norms are negatively associated with prediction confidence. In the region of low norms, there is a linear relation between the norm and the prediction entropy, and this can be further used also to reject samples at classification time.

\section{Experiments}
\label{sec:results}
We first evaluate the learned representations on cropped\footnote{Using the biased~\cite{Kumar:ICCV09, DBLP:journals/corr/SunWT14} background to improve performance is not in the scope of this work.} faces of the Labeled Faces in the Wild (LFW) public dataset~\cite{Huang:ECCVW08}, using multiple protocols. We also validate our findings on an internal dataset, probing 100K faces among 10K subjects with a probe-gallery identification protocol.

The LFW dataset consists of 13,233 web photos of 5,749 celebrities, and is commonly used for benchmarking face verification.
In this work, we focus on two new Probe-Gallery \emph{unsupervised} protocols proposed in
~\cite{Lacey:MSUTR} (the original splits are used):
\begin{enumerate} 

\item A closed set identification task, where the gallery set includes 4,249 identities, each with only a single example, and the probe set includes 3,143 faces belonging to the same set of identities. The performance is measured by the Rank-1 identification accuracy.
\item An open set identification task, where not all probe faces have a true mate in the gallery. The gallery includes 596 identities, each with a single example, and the probe set includes 596 genuine probes and 9,491 impostor ones. Here the performance is measured by the Rank-1 Detection and Identification Rate (DIR), which is the fraction of genuine probes matched correctly in Rank-1 at a 1\% False Alarm Rate (FAR) of impostor probes that are not rejected.
\end{enumerate}

As the verification protocol, we follow the LFW unrestricted protocol~\cite{newlfw} (which uses only the same-not-same labels), and similarly to~\cite{Taigman:CVPR14} train a kernel SVM (with C=1) on top of the $\chi^2$-distance vectors derived from the computed representations. For the open and closed set identification experiments, we simply use the cosine similarity (normalized dot product). \textbf{A critical difference} between the LFW verification protocols and the Probe-Gallery ones is that the latter does not permit training on the LFW dataset. They, therefore, demonstrate how face recognition algorithms perform on an unseen data distribution, as close as possible to real-life applications. Also, as pointed out in~\cite{Lacey:MSUTR}, the Probe-Gallery protocols correspond to many challenging practical scenarios, such as retrieval~\cite{Klontz:Boston13}. The importance of such protocols as tools that differentiate face recognition methods based on performance is confirmed by our results, since methods that exhibit very similar results on the LFW verification protocol display large performance gaps on the Probe-Gallery ones, as shown in Table~\ref{table-dim}. 
In all of our experiments we follow the unrestricted protocol using labeled outside data~\cite{newlfw}. We reiterate the importance of having a large scale training set in hand: training the initial net~\cite{Taigman:CVPR14} using 500K images on CASIA~\cite{Yi:arXiv14}, the largest public face dataset available today, obtains verification accuracy of only 93.5\% on LFW, largely due to over-fitting. The code for training the nets is written in Torch and uses the fbcunn extensions (see: ~\cite{fbcunn}). 

\subsection{Compressed Representations}
\label{sec:crexp}
We first evaluate different compressed representations, all utilizing the initial face representation system, with sizes ranging from 4096 dimensions down to 8. These networks were retrained as described in Sec.~\ref{sec:compress} on the 4 million images associated with 4,030 random identities used in~\cite{Taigman:CVPR14}.

Table~\ref{table-dim} shows that compression improves generalization considerably. With only 256 dimensions, the obtained Rank-1 accuracy stands on 72.3\% on the Closed Set protocol, and DIR 46.3\% at 1\% FAR on the Open Set, and greatly outperforms the original 4096D representation. Note, however, that the difference in the verification protocol remains within a 1\% range when compressing the 4096 dimensions down to 64, and either due to performance saturation and/or the type of benchmark, differences in face recognition capabilities are not captured well.

\begin{figure*}[t]
\centering
\includegraphics[width=0.425\linewidth,clip,trim=1.8cm 0.7cm 1.2cm 0.2cm]{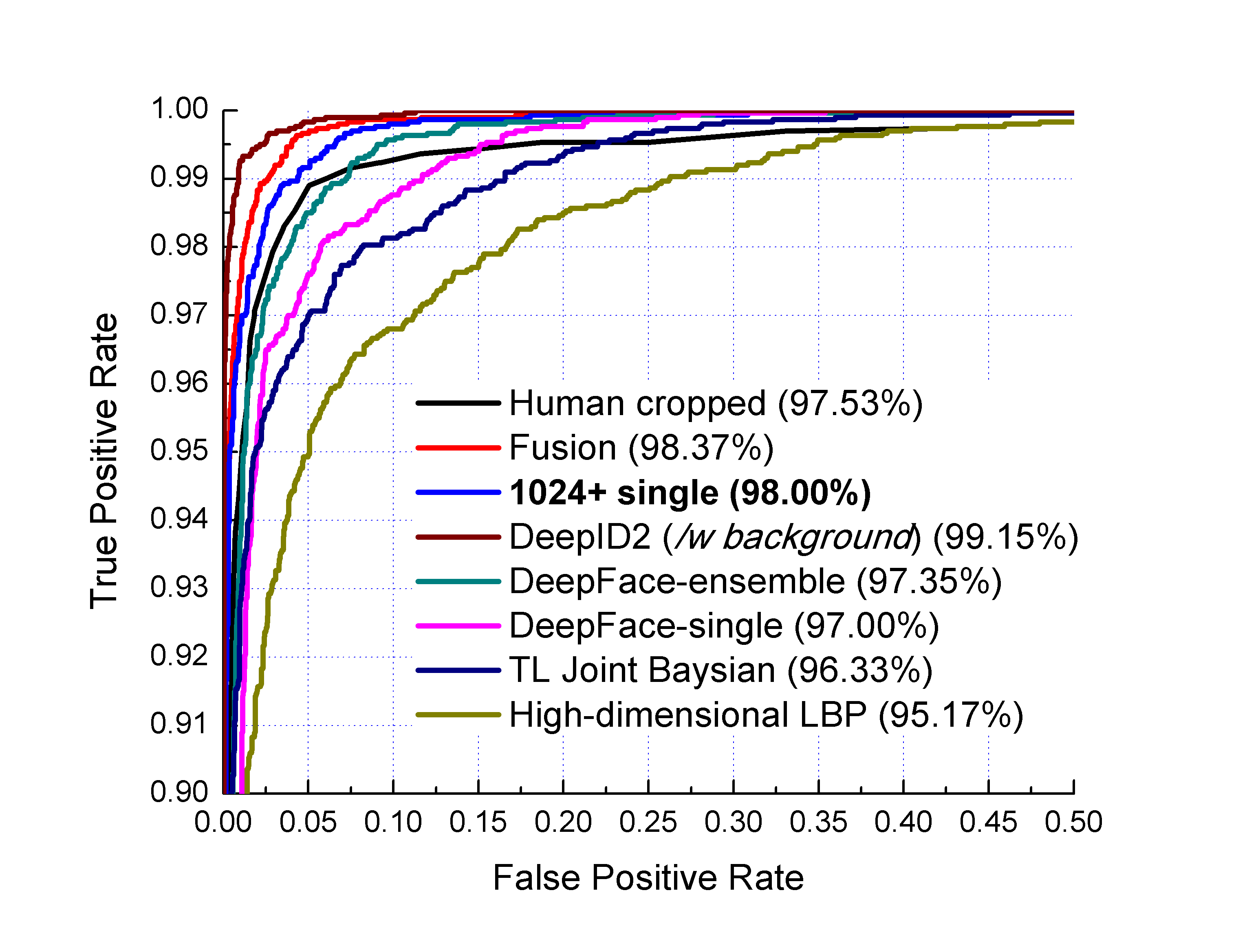}
\includegraphics[width=0.425\linewidth,clip,trim=1.8cm 0.7cm 1.2cm 0.2cm]{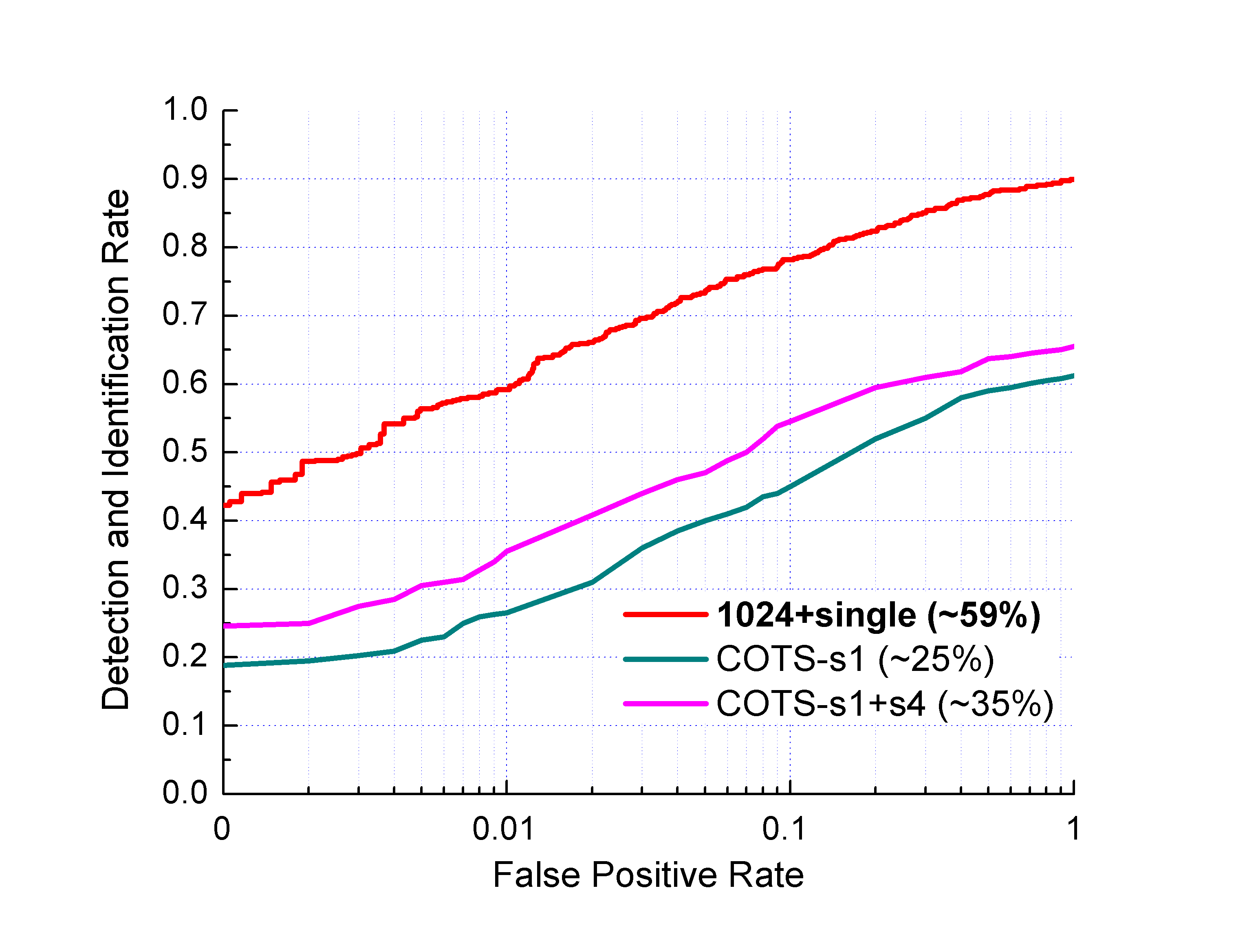}
\caption{\textbf{Left:} The ROC curves on the face verification unrestricted protocol. \textbf{Right:} The DIR vs. FAR curves on the Open Set protocol. As mentioned above, for identification, training on LFW images is not permitted as it invalidates the comparison to baselines; had we jointly fit a multi-class linear SVM to the gallery, the best model would achieve 69\% DIR @ 1\% on the open set. Best viewed in color. The ordinate scales are different. COTS graphs were reconstructed from~\cite{Lacey:MSUTR}.}
\label{fig:curve_lfw}
\end{figure*}

\begin{table}[t]
\centering
\begin{small}
\begin{tabular}{|@{~}l@{~}|@{~}l@{~}l@{~}l@{~}l@{~}l@{~}l@{~}l@{~}l@{~}l@{~}@{~}|}
\hline
Dim. & 4096 & 1024 & 512 & \bf{256} & 128 & 64 & 32 & 16 & 8 \\
\hline\hline
Verifica. & 97.00 & 96.72 & 96.78 & \bf{97.17} & 96.42 & 96.10 & 94.50 & 92.75 & 89.42 \\
\hline
Rank-1        & 60.9  & 64.9  & 67.4  & \bf{72.3}  & 69.1 & 66.5  & 39.6  & 23.2  & 7.00 \\
Rank-10       & 78.7  & 83.9 & 85.2 & \bf{90.4} &  88.8 & 87.7 & 70.8 & 52.9 & 24.7 \\
DIR@1\% & 41.9  & 44.7 & 46.1 & \bf{46.3} & 44.1 & 36.7 & 12.2 & 5.37& 0.33 \\
\hline
\end{tabular}
\end{small}
\caption{Performance on the three protocols, when varying the dimensionality of the representations. Performance is measured in terms of the verification accuracy (\%) of the \emph{unrestricted} verification protocol, Rank-1 and Rank-10 accuracy (\%) on the Closed Set, and the DIR (\%) at 1\% FAR on the Open Set. }
\label{table-dim}
\end{table}

\subsection{Bootstrapped Representations}
We now compare the representation learned on the 55K bootstrapped identities (4.5M faces) with those learned from randomly selected 108K identities (3.2M faces) and even 250K identities (7.5M faces), as shown in Table~\ref{table-dataset}.
We note that: (i) The deep neural network (DNN) can benefit from additional amount of training data, \emph{e.g.}, 250K identities, boost the recognition performance over those trained on 4K identities in Table~\ref{table-dim}. (ii) The bootstrapped training set of 55K identities, although five times smaller than the biggest training set used, delivers better Probe-Gallery performance. (iii) An even larger improvement is obtained when the locally-connected layers (L4-L5-L6) are expanded as described in Sec.~\ref{sec:32filters}, and the extended 256D and 1024D representations (denoted as 256+ and 1024+) generalize better than their unmodified counterparts. Larger networks were also attempted but failed to improve performance, e.g. 2048+ reduced Rank-1 accuracy by 4.21\% on the closed set.

\begin{table}[t]
\centering
\begin{small}
\begin{tabular}{|l|llll|}
\hline
 & \multicolumn{4}{c|}{Random 108K}  \\
 & 256 & 512 & 1024 & 2048   \\
\hline\hline
Verifca. & 97.35 & 97.62 & 96.90 & 96.47\\
\hline
Rank-1  & 69.7 & 68.1 & 70.2 & 68.4 \\
Rank-10 & 88.5 & 87.8 & 90.1 & 88.1 \\
DIR@1   & 51.3 & 46.5 & 51.0 & 47.0 \\
\hline
& \multicolumn{4}{c|}{Random 250K} \\
&  256 & 512 & 1024 & 2048 \\
\hline
\hline
Verifca.& 96.33 & 97.10 & 97.67 & 96.30 \\
\hline
Rank-1 & 59.6 & 74.0 & 74.9 & 63.9 \\
Rank-10 & 86.0 & 91.7 & 90.9 & 81.7 \\
DIR@1 & 38.1 & 54.7 & 58.7 & 45.3 \\
\hline
& \multicolumn{4}{c|}{Bootstrap 55K} \\
& 1024 & 256+ & 1024+ & 2048+ \\
\hline
\hline
Verifca. & 97.57 & 97.58  & \bf{98.00} &  97.92 \\
\hline
Rank-1 & 75.9 & 77.0  & \bf{82.1}  & 77.9 \\
Rank-10 & 91.1 & 91.7  & \bf{93.7} & 92.0 \\
DIR@1 & 56.2  & 57.6  & \bf{59.2}  & 54.5 \\
\hline
\end{tabular}
\end{small}
\caption{Performance of the three protocols on different training sets. The three rightmost columns (denoted by 256+, 1024+, and 2048+) report results using the architecture discussed in Sec.~\ref{sec:32filters}.}
\label{table-dataset}
\end{table}

\subsection{Comparison with the State-of-the-art}

The state of the art COTS face recognition system, as evaluated by NIST in~\cite{Grother:MBE10}, and diligently benchmarked by~\cite{Lacey:MSUTR} on the LFW open and closed set protocols provides a unique insight as to how well our system compares to the best commercial system available.
The authors of~\cite{Lacey:MSUTR} have also employed an additional vendor that rectifies non-frontal images by employing 3D face modeling and improved the results of the baseline \textbf{COTS-s1} system, the combined system is denoted \textbf{COTS-s1+s4}. For further comparison, we have evaluated the publicly available LFW high dimensional LBP features of~\cite{Chen:CVPR13}, denoted as \textbf{BLS}, which are published online in their raw format, i.e. \emph{before} applying the prescribed supervised metric learning method. Finally, in order to push our results further, we fuse four of the networks trained in our experiments (namely, the initial, Random-108K, Random-250K and 1024+ Bootstrap-55K, all 1024d) by simply concatenating their features, and report a slight improvement, denoted as \textbf{Fusion}. Note that for the LFW verification benchmark only, \cite{DBLP:journals/corr/SunWT14} reports an accuracy of 99.15\%. In contrast to our work, this result employs hundreds of CNNs fused with background information from the LFW images. Our network is limited to the face region (`cropped') which helps remove important dataset-specific biases~\cite{Kumar:ICCV09} and even with a single network we achieve 98.0\% verification performance. For instance, a single network of~\cite{DBLP:journals/corr/SunWT14} obtains up to 95.43\%.
Table~\ref{table-compare} and Figure~\ref{fig:curve_lfw} summarize these results.
Our best method lowers the state of the art miss rate on the closed set protocol by 57\%, and by 45\%, at the same precision level, on the open set protocol. The error in the verification protocol is reduced by 38\% with respect to the initial baseline system.
\begin{table}[t]
\centering
\begin{small}
\begin{tabular}{|l|l|l|l|l|l|l|}
\hline
Method & DF & BLS & COTS$\!\!$ & COTS$\!\!$ & \bf{1024+}$\!\!$ &  \bf{Fusion}$\!\!$ \\
& {\small ~\cite{Taigman:CVPR14}} &  {\small \cite{Chen:CVPR13}$^*$} & {\small s1~\cite{Lacey:MSUTR}$\!$} & {\small s1+s4~\cite{Lacey:MSUTR}$\!\!$} & & \\
\hline\hline
Verifica.    & 97.35 & 93.18 & - & - & 98.00 &  98.37\\
\hline
Rank-1          & 64.9  & 18.1  & 56.7 & 66.5 & 82.1 & 82.5  \\
DIR @ 1\%$\!\!$ & 44.5  & 7.89  & 25 & 35 & 59.2 & 61.9 \\
\hline
\end{tabular}
\end{small}
\caption{Comparison to state of the art that includes the COTS method and two recent methods, in terms of the Probe-Gallery's Rank-1 accuracy (\%) on the Closed Set, the DIR at 1\% FAR on the Open Set, as well as the verification protocol. $^*$For~ \cite{Chen:CVPR13} only the published raw features are used and not the full system. The full system achieves 95.17\% on the verification task. For both COTS the verification performance was not reported.}
\label{table-compare}
\end{table}

\begin{figure}[t]
\centering
\includegraphics[width=1.0\linewidth]{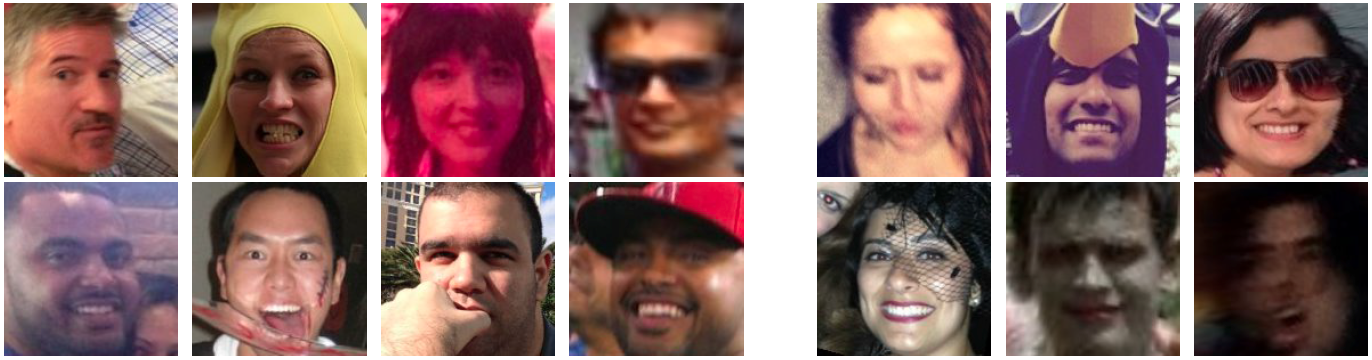}
\caption{Examples of successful (left) and failed (right) probed identifications on our validation set. A successful match is when a probe achieves the highest similarity to its true-mate in the gallery (Rank-1), failure otherwise. Permission granted by the subjects.}
\label{fig:examples}
\end{figure}

\subsection{Model Selection}
\label{sec:modelsec}
Since the LFW verification protocol~\cite{Lacey:MSUTR} does not allow model selection of any kind, we selected the best model using a separate internal validation dataset, that consists of 10,000 individuals. As gallery we take 55,000 images (an average of 5.5 face images, per identity). At test time, we perform 10 queries per person, using an additional set of 100,000 images.
We search the probe faces with the cosine similarity using the 4096 dimensional initial representation in~\cite{Taigman:CVPR14}, the compressed 256D and the bootstrapped 1024+ representations.
Performance is measured by the Rank-1 accuracy and the DIR at 1\% and 0.1\% FAR.
The results are listed in Table~\ref{table-social}, and confirm that (i) the compressed 256D representation generalizes better than the 4096D one; and (ii) the bootstrapped 1024+ improves the DIR substantially for the low 0.1\% FAR. A few examples of successful and failed probed faces are shown in Fig.~\ref{fig:examples}.
\vspace{-4.45mm}
\paragraph{Representation norm in retrieval.} We have also verified on this $10,000$ identity validation set that the representation norm is linked to the identification accuracy. As shown in Fig.~\ref{fig:norms}(d), there's a clear correlation between the retrieval success, measured in terms of the minimum rank of one of the true gallery images in the retrieval list, and the representation norm. The correlation is extremely strong ($\rho$ = -0.251), and there is a sizable gap between the mean norm of the successful rank-1 queries and the unsuccessful ones.

\begin{table}[t]
\centering
\begin{small}
\begin{tabular}{|l|c|c|c|}
\hline
 & 4096 & 256 & \bf{1024+}  \\
\hline\hline
Rank-1       & 67.87  &  70.60  &  72.94 \\
DIR @ 1\%    & 51.02  &  53.39  &  59.60 \\
DIR @ 0.1\%  & 37.79  &  38.96  &  46.97 \\
\hline
\end{tabular}
\end{small}
\caption{Probe-Gallery results on an internal dataset used for model selection.}
\label{table-social}
\end{table}

\section{Summary}
\vspace{-1.45mm}
Face recognition is unlike any other recognition task in several ways. First, although it is only one object, it is by far the most frequent entity in the media, and there are billions of unique instances (identities) to differentiate between. Second, since the universe of faces is open in practice, the most interesting problems are Transfer Learning tasks, where it is required to learn how to represent faces in general. This representation is then tested on unseen identities. This is in contrast to training on closed tasks such as ImageNet's Image Classification challenge where recognition performance is optimized directly on a fixed set of 1000 classes.
Accurate face alignment enables us to concentrate solely on the underlying inter-personal variances that exist in face recognition using deep convolutional nets. Recent works in the domain of face recognition have yielded impressive results by utilizing large and deep convolutional nets. However, there is no clear understanding of why they perform so well and what the important factors in such systems are. 
We explore the task of transferring representation of faces in a number ways. First, we identify and explain the role of the network's bottleneck as an important regularizer between training-set specificity and generality. Second, we have identified a saturation point in performance, as the number of training samples grows beyond what has been explored in the past, and provided an efficient method that alleviates this by modifying the common practice of randomly subsampling the training set. Third, we show how the representation layer of faces is constructed and affected by distortions, linking between its reduced norm to the undesirable increase in uncertainty.
Lastly, our work is unique in that it allows a direct comparison of commercial systems to those published in the academic literature, on the benchmark task used in practice to compare actual commercial systems. We conjecture that the impact of these discoveries goes beyond face recognition, and is applicable to other large scale learning tasks. 

\newpage

{
\bibliographystyle{ieee}
\bibliography{IEEEabrv,cvpr2015}
}

\end{document}